\documentclass{bmvc2k}

\usepackage{times}
\usepackage{epsfig}
\usepackage{graphicx}
\usepackage{amsmath}
\usepackage{amssymb}
\usepackage{bm}
\usepackage{booktabs,colortbl,tabularx}
\usepackage{footmisc}

\usepackage{mathtools}
\usepackage{commath}

\usepackage{algorithm}
\usepackage[noend]{algpseudocode}
\usepackage{bbm}
\usepackage{comment}
\usepackage{multirow}
\usepackage{enumitem}

\usepackage{pifont}

\newcommand{\cmark}{\ding{51}}%
\newcommand{\xmark}{\ding{55}}%

\newcommand{\vect}[1]{\bold{\boldsymbol{#1}}}

\definecolor{battleshipgrey}{rgb}{0.52, 0.52, 0.51}

\let\norm\undefined 
\DeclarePairedDelimiter\norm{\lVert}{\rVert}

\def\thefootnote{*}\footnotetext{Corresponding author.}

\title{Siamese Prototypical Contrastive Learning}

\addauthor{Shentong Mo}{shentonm@andrew.cmu.edu}{1}
\addauthor{Zhun Sun\thefootnote{}}{sun@vision.is.tohoku.ac.jp}{2}
\addauthor{Chao Li}{chao.li@riken.jp}{3}
\addinstitution{
 Carnegie Mellon University\\
 Pittsburgh, PA 15213, United States
}
\addinstitution{
 Tohoku University\\
 Sendai, Miyagi, Japan
}
\addinstitution{
 Center for Advanced Intelligence Project (AIP), RIKEN \\
 Tokyo, Japan
}

\runninghead{Shentong Mo, Zhun Sun, and Chao Li}{SPCL}

\begin{document}

\maketitle

\begin{abstract}
    Contrastive Self-supervised Learning (CSL) is a practical solution that learns meaningful visual representations from massive data in an unsupervised approach.
    The ordinary CSL embeds the features extracted from neural networks onto specific topological structures. During the training progress, the contrastive loss draws the different views of the same input together while pushing the embeddings from different inputs apart.
    One of the drawbacks of CSL is that the loss term requires a large number of negative samples to provide better mutual information bound ideally.
    However, increasing the number of negative samples by larger running batch size also enhances the effects of false negatives: semantically similar samples are pushed apart from the anchor, hence downgrading downstream performance.
    In this paper, we tackle this problem by introducing a simple but effective contrastive learning framework. The key insight is to employ siamese-style metric loss to match intra-prototype features, while increasing the distance between inter-prototype features.
    We conduct extensive experiments on various benchmarks where the results demonstrate the effectiveness of our method on improving the quality of visual representations. Specifically, our unsupervised pre-trained ResNet-50 with a linear probe, out-performs the fully-supervised trained version on the ImageNet-1K dataset.
\end{abstract}
\vspace{-1em}
\section{Introduction}
\vspace{-0.5em}
Massive studies have recently shown that Contrastive Self-supervised Learning (CSL) is a powerful tool for unsupervised pre-training models and semi-supervised training methods~\cite{caron2018deep,caron2019unsupervised,caron2020unsupervised,chen2020simple,chen2020mocov2,he2019moco,li2021prototypical}. 
The core idea of CSL is to utilize the views of samples to construct a discrimination pretext task. 
This task learns the capability of the deep neural network extracting meaningful feature representations, which can be further used by tons of downstream tasks, such as image classification, object detection, and instance segmentation.

This idea of CSL has appeared in former literature a long time ago~\cite{dosovitskiy2014discriminative,hadsell2006dimensionality,virginia1993learning,suzanna1992self-organizing,peter1991learning}. 
The term ``view'' in CSL generally refers to augmented samples, and the goal of CSL is to close the distance between different views from the same sample, \textit{a.k.a.} the anchor, while keeping views from different samples away from the anchor. 
This goal is commonly achieved in recent studies by employing the InfoNCE~\cite{oord2018representation} loss and its variants. 
The ideal outcome of a contrastively pre-trained model should distribute the samples in the embedding space uniformly by maximizing the mutual information between the anchor and its views.

However, theoretically, it is demonstrated that the loss term requires a large number of negative samples to provide better mutual information bound~\cite{poole2019variational,tschannen2019mutual}, which leads to large batch sizes. 
Although the growth of computation resources nowadays can support training with batch sizes of dozens of thousands, the large mini-batch often results in new trouble: The chance of semantically similar samples being computed as negative ones is raised. 
To push apart the \textit{false negative} pairs in the embedding space does not help to build a semantic understanding of the samples, and does actually damage the performance of the pre-trained model when the downstream tasks heavily rely on the semantic information~\cite{arora2019theoretical,tschannen2019mutual}.

\begin{figure}[t]
\setlength{\abovecaptionskip}{-1.5em}
\setlength{\belowcaptionskip}{-1em}
	\centering
		\includegraphics[width=0.6\linewidth]{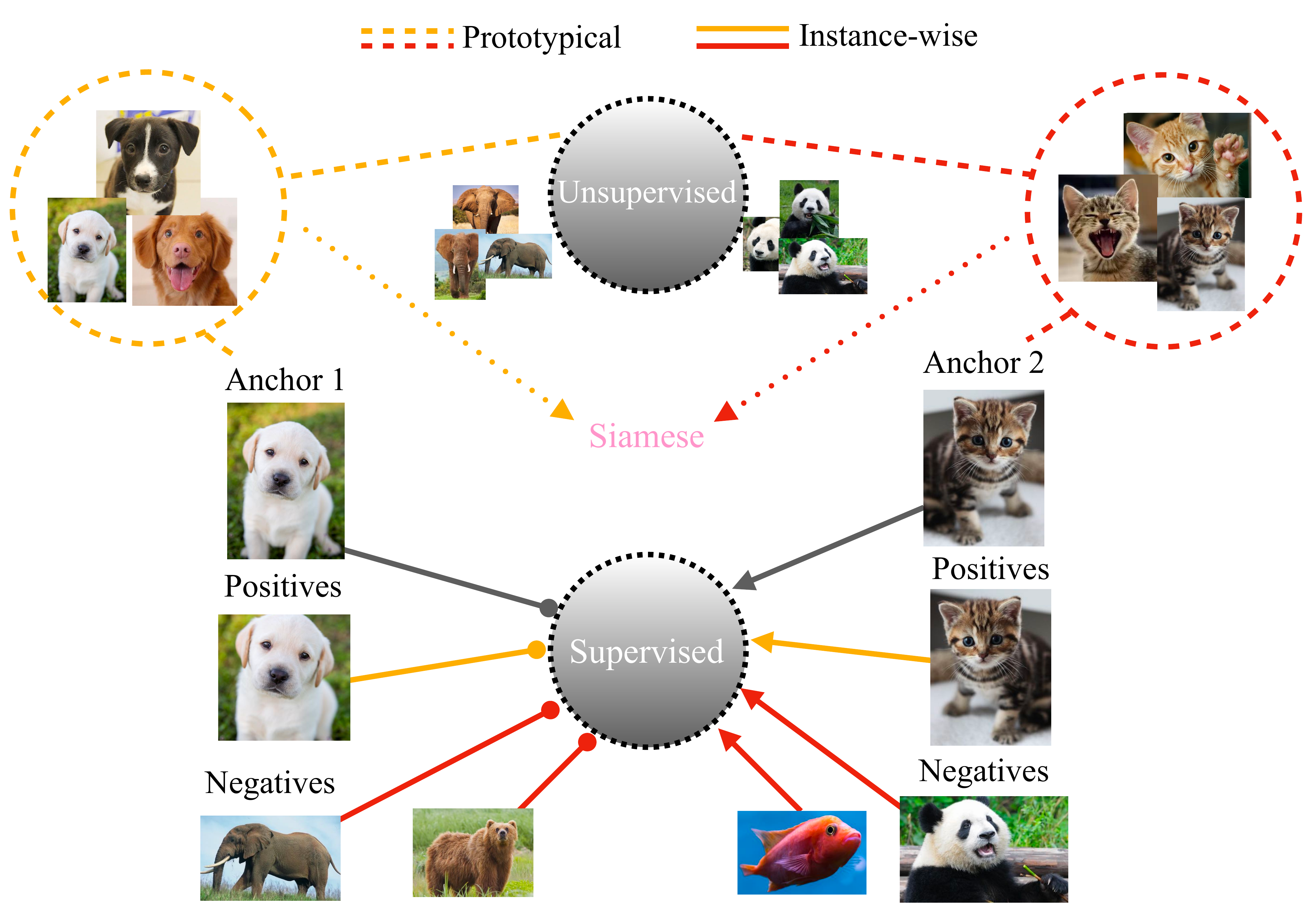}    	\caption{Illustration of our Siamese Prototypical Contrastive Learning. The embedded features are grouped into separate prototypes, and a Siamese-style metric loss is employed to match intra-prototype while pushing apart inter-prototype features. Instance-wise contrastive learning is conducted inside each separate prototype. And the views of the anchors are regularized to the corresponding prototypes.
    	}
    	\vspace{-0.2cm}
	\label{fig:title_img}
\end{figure}

One straightforward approach to address it is to employ the label information of the samples explicitly, \textit{i.e.} supervised contrastive learning~\cite{khosla2020sup}. The samples are regularized to the corresponding semantic clusters with an ordinary cross-entropy loss. 
The obvious shortcoming of this approach is the prerequisite of the downstream tasks knowledge. 
Another approach is to use "prototypes" to tackle the issue in a task-agnostic manner.
The term ``prototype'' refers to the clusters of the samples in the embedding space. 
The clusters can be either pre-defined or computed during training mini-batches. 
The samples from the same prototype contain toughly similar semantic information. 
Thus, the learned embedded features provide better discriminability for the downstream classification tasks. 

In this work, we follow the instance-wise contrastive learning but also regularize the views of samples with supervised prototypes. 
In our initial experiments, we find that a straightforward addition of the prototypes' cross-entropy loss decreases the performance. 
Moreover, it is not helpful to separate the cluster centers by minimizing their similarity. 
We conjecture that it is not feasible to distinguish the semantic false-negative samples in the original embedding space. 
To address this problem, we propose a Siamese-style metric loss~\cite{koch2015siamese} to maximize the inter-prototype distance while preserving the intra-prototype similarity.

From our experimental analysis, we empirically reveal that the similarity of the false-negative samples to their anchors decreases in the contrastive projection space. We present the overall framework as Siamese Prototypical Contrastive Learning, the main contributions of this paper can be summarized as follows:
\vspace{-0.5em}
\begin{itemize}[leftmargin=*]
    \item We propose a simple but effective contrastive learning framework, utilizing the Siamese-style metric loss to improve the semantic discriminability between prototypes.
    \vspace{-0.5em}
    \item We show that our SPCL successfully mitigates the confusion brought by false negatives.
    \vspace{-0.5em}
    \item Our SPCL outperforms its baseline by a large margin and achieves state-of-the-art performances on the ImageNet-1K dataset. 
\end{itemize}

\vspace{-1em}
\section{Related Works}

\vspace{-0.5em}
\subsection{Self-supervised Learning}
\vspace{-0.5em}
In recent years, self-supervised learning methods have gained popularity in learning visual representations. 
DeepCluster~\cite{caron2018deep} performs iterative clustering and unsupervised representation learning, which is further improved by DeeperCluster~\cite{caron2019unsupervised} to learn transferable representations from a large scale of images and online clustering~\cite{zhan2020online}.
In this work, we intend to employ clustering to boost the performance of contrastive self-supervised learning of visual representations. 
The objective functions used to train visual representations in many methods are either reconstruction-based loss functions or contrastive loss that measures the co-occurrence of multiple views~\cite{caron2020unsupervised,chen2020simple,chen2020big,chen2020mocov2,he2019moco,tian2020contrastive}. 
The effectiveness of self-supervised learning in visual representation can be attributed to two important aspects, namely pretext tasks and contrastive learning. 
For pretext tasks, there has been a wide range of pretext tasks explored to learn
a good representation. These examples include colorization~\cite{zhang2016colorful,yamaguchi2019multiple}, context autoencoders~\cite{deepak2016context}, spatial jigsaw puzzles~\cite{misra2020self,noroozi2016unsupervised} and discriminate orientation~\cite{xiao2020contrastive}.
For contrastive learning, that approaches~\cite {grill2020bootstrap} bring the representation of different views of the same image (positive pairs) closer, and
push representations of views from different images (negative pairs) apart using instance-level classification with
contrastive loss.

\vspace{-1em}
\subsection{Contrastive Learning}
\vspace{-0.5em}

Recently, researchers have been interested in exploring effective methods using contrastive self-supervised learning. 
The breakthrough approach is SimCLR~\cite{chen2020simple}, which is trained by an end-to-end structure, where the features of each instance are pulled away from those of all other instances in the training set. In-variances are encoded from low-level image transformations such as cropping, scaling, and color jittering. This sort of end-to-end structures~\cite{chen2020simple,chen2020big} always use large batch size to accumulate a large bunch of negative samples in a contrastive loss.  
Instead of using a large batch size, PIRL~\cite{misra2020self} applies a memory bank to store and update representations learned at a specified stage, which would be computationally expensive. 
To update the stored memory efficiently, MoCo~\cite{he2019moco} and MoCov2~\cite{chen2020mocov2} replace the memory bank with a memory encoder to queue new batch samples and to dequeue the oldest batch. 
A concurrent work~\cite{chen2021simsiam} explores a simple Siamese network to maximize the similarity between two views of the anchor. DenseCL~\cite{wang2020dense} and SCRL~\cite{byungseok2021spatial} apply contrastive learning on the pixel-level and spatial-level features to boost the performance of tasks related to images.

\vspace{-1em}
\subsection{Prototypical Contrastive Learning}
\vspace{-0.5em}

Compared to instance-wise contrastive methods, relatively few works consider using prototypical contrastive methods. More recent work, SwAV~\cite{caron2020unsupervised} focuses on using a ``swapped'' clustering-based mechanism to group similar representations, which enforce consistency between
cluster assignments produced for different views of the anchor. PCL~\cite{li2021prototypical} uses prototypes as latent variables and performs iterative clustering in an expectation-maximization (EM)-based framework using the proposed ProtoNCE loss. In this work, we employ clustering as an unsupervised learning method to generate prototypes as a ``pseudo label''. Furthermore, we propose the siamese-style metric loss to match the intra-class features and increase the distance between inter-class features. Our ``metric'' loss is similar to the supervised contrastive loss in SupCon~\cite{khosla2020sup}.
However, we do not use the ground-truth directly as supervised information and instead leverage the clustering to generate ``pseudo label''.

\setlength{\belowdisplayskip}{0.7pt} \setlength{\belowdisplayshortskip}{0.7pt}
\setlength{\abovedisplayskip}{0.7pt} \setlength{\abovedisplayshortskip}{0.7pt}

\begin{algorithm}[t]
\caption{SPCL main learning algorithm}
\label{alg:main}
\begin{algorithmic}[1]
    \Statex {\bfseries Input:} Data $X$, number of prototypes $N$, $f, g_{c}, g_{m}, g_{p}$, sets of augmentation $\mathcal{T}$.
    
    \For{each epoch}
    \State Re-initialize $g_{c}$, $g_{m}$, $g_{p}$.
    \State Obtain clusters of prototypes $\boldsymbol{C} = \{\boldsymbol{c}^0, ...,\boldsymbol{c}^n\}$ using $K$-means clustering.

    \For{each step}
    
    \State Choose the prototype $p, q$.
    \State Sample $\bm{x}^{p}$, $\bm{x}^{q}$ from the corresponding clusters $\boldsymbol{c}^p$, $\boldsymbol{c}^q$.
    
    \State Obtain $\bm{h}^{p}$, $\bm{h}^{q}$ from the inputs and their augmented views.

    \State Compute the $L_{contra}, L_{metric}, L_{proto}$ using Eq.~\ref{eq:csl2},~\ref{eq:mel},~\ref{eq:cel}.
    
    \State Compute $L_{total}$ in Eq.~\ref{total_loss}, and update networks $f, g_{c}, g_{m}, g_{p}$ to minimize $L_{total}$
    
    \EndFor
    
    \EndFor

\end{algorithmic}
\vspace{-0.5em}
\end{algorithm}

\section{Method}\label{sec:method}
\vspace{-0.5em}
We first illustrate the main framework of our proposed Siamese-style Prototypical Contrastive Learning (SPCL) and then provide technical details in the following subsections. 

\subsection{Overview and Notation}
\vspace{-0.5em}
\noindent{\textbf{Overview.}} Our goal is to learn meaningful visual representations in an unsupervised and task-agnostic style. 
To achieve this, we present a simple but effective variant of contrastive self-supervised learning. 
The fundamental idea of this method is to group the samples in the embedding space to produce prototypes; then, we refine the prototypes by maximizing/minimizing the inter-/intra-cluster distance between samples. Finally, we regularize the views of samples using the prototypical information with cross-entropy.

Our SPCL contains four major modules: the start-epoch clustering, the ordinary contrastive loss, the Siamese-style metric loss, and the prototypical cross-entropy loss, which we will introduce below. Fig.~\ref{fig:title_img} and Algo.~\ref{alg:main} provide a rough illustration of our method. 

\noindent{\textbf{Notation.}} 
We use italic letters to denote scalars, \textit{e.g.}, $n,N\in{}\mathbb{R}$, and use boldface letters to denote vectors/matrices/tensors, \textit{e.g.}, $\bm{h,y}\in{}\mathbb{R}^{n}$
and $\bm{x}\in\mathbb{R}^{h\times{}w\times{}c}$. 
We denote \textit{sets} by calligraphic letters, \textit{e.g.} $\mathcal{X},\mathcal{T}$ stands for the set of samples and transformations, correspondingly.
Sometimes italic letters are also used to represent functions or procedures, we employ the form $f(\cdot)$ to exclude ambiguity.

\subsection{Clustering and Feature Extracting}
\vspace{-0.5em}
\noindent{\textbf{Clustering.}} 
The first step of our SPCL is to group the embedded features into separate clusters in an unsupervised manner. 
This is achieved by the simple $K-$means algorithm, where the number of prototypes $K$ is a preset hyper-parameter. 
Concretely, at the beginning of every epoch, we split the dataset $\mathcal{X}$ into mini-batches. For each mini-batch, we draw an augmentation $t\sim \mathcal{T}$ and extract the features of the augmented samples. 
We concatenate all the embedded features and conduct the clustering algorithm to obtain the set of prototypes $\mathcal{C} = {\bm{c}^1, \ldots, \bm{c}^K}$, where $\bm{c}^k$ is the vector of index recording samples of the cluster $k$. 

\noindent{\textbf{Feature Extracting.}} During the feature extracting phase,
we first draw an anchor prototype $p$ and sample a mini-batch $\mathcal{X}^{p} = \{\bm{x}_n^p\}^N_{n=1}$ from $\bm{c}^p$. 
Then, we draw samples that do not belong to prototype $p$, \textit{i.e.} from $\mathcal{C} \setminus \{\bm{c}^p\}$. 
and denote this mini-batch with $\mathcal{X}^{q} = \{\bm{x}_n^q\}^N_{n=1}$, where $q$ stands for any prototypes that are not $p$. 
For each sample inn $\mathcal{X}^{p}$ and $\mathcal{X}^{q}$, we acquire its \textit{two} augmented views and the feature representations in the embedding space. 
The feature representations are denoted by $\mathcal{H}^p = \{ \bm{h}_{n,1}^p, \bm{h}_{n,2}^p\}^N_{n=1}, \mathcal{H}^q = \{ \bm{h}_{n,1}^q, \bm{h}_{n,2}^q\}^N_{n=1}$, with all their elements $\bm{h} \in \mathbb{R}^{d_e}$, where $d_e$ is the dimension of the embedding space.

\subsection{Contrastive loss}
\vspace{-0.5em}
We employ the ordinary \textbf{NT-Xent} (the normalized temperature-scaled cross entropy loss)~\cite{sohn2016improved,  wu2018unsupervised,oord2018representation} as our contrastive loss. 
Following SimCLR~\cite{chen2020simple}, we use a projection head $g_c(\cdot)$ that maps the extracted representations $\bm{h}$ to the space where contrastive loss is applied. 
That is, $\bm{z} = g_c(\bm{h})$. 
We use $s(\bm{u}, \bm{v}) = \vect{u}^T\vect{v} / \norm{\vect{u}}\norm{\vect{v}}$ to denote the dot product between $\ell_2$ normalized $\vect{u}$ and $\vect{v}$ (\textit{i.e.} cosine similarity). Then the contrastive loss of a sample $\bm{z}$ is defined as  
\begin{equation}\label{eq:csl1}
    L_{contra, \bm{z}} = -\log 
    \frac{\exp(s(\bm{z}, \bm{z}') /\tau)}
    {
    \sum\limits_{\substack{\bar{\bm{z}} \in \{\mathcal{Z}^p\cup \mathcal{Z}^q\}, \\ \bar{\bm{z}} \neq \bm{z}, \bm{z}'}}
    \exp(
        s(\bm{z}, \bar{\bm{z}})/\tau)
        )
    },
\end{equation}
where $\mathcal{Z}$ is the set of representations in the projection space, $\bm{z}'$ is the representation of the other view of the sample, $\tau$ denotes a temperature parameter. The final contrastive loss is computed as the sum-reduce across all samples:
\begin{equation}\label{eq:csl2}
    L_{contra} = 
    \sum\limits_{\bm{z} \in \{\mathcal{Z}^p\cup \mathcal{Z}^q\}}
    L_{contra, \bm{z}}
\end{equation}

\vspace{-1em}
\subsection{Siamese-style metric loss}
\vspace{-0.5em}
In our initial experiments, we observe that the prototypical cross-entropy loss employment does not improve the performance of the downstream tasks. One reason is that the clusters of prototypes are not well separated; the samples that are far away from the cluster centers might be \textit{false positive} and are aggressively pulled towards the anchor sample. 
As a consequence, the semantic information of the prototypes becomes more obscure. 
To tackle this problem, we introduce a Siamese-style metric loss, which matches intra-class features and increases the distance between inter-class features. Concretely, for each feature representation of the anchor prototype $p$, we draw two feature embeddings from $p$ and $q$ as positive and negative pairs, denoted as $\{\bm{h}^p, \hat{\bm{h}}^p\}$ and $\{\bm{h}^p, \hat{\bm{h}}^q\}$. 
It is worth mentioning that, here, we still slightly abuse the notation $q$, which stands for \emph{a prototype that is not $p$}. 
With a projection head $g_{m}(\cdot)$ that predicts the distance between the pairs, the Siamese-style metric loss can be written as
\begin{equation}\label{eq:mel}
    L_{metric} = \sum\limits_{\bm{h}^p \in \mathcal{H}^p} 
    \Big(\mathtt{CE}(g_m(\vert \bm{h}^p - \hat{\bm{h}}^p \vert|, 1) +
    \mathtt{CE}(g_m(\vert \bm{h}^p - \hat{\bm{h}}^q \vert|, 0) \Big)
\end{equation}

\vspace{-1em}

\subsection{Prototypical cross-entropy loss}
\vspace{-0.5em}
Motivated by SupCon~\cite{khosla2020sup}, we introduce the prototypical cross-entropy loss in our SPCL to guide the views of one sample to its corresponding prototype. This loss term is helpful when the similarity between views is small, \textit{i.e.} $s(\bm{z}, \bm{z}') \approx 0$, and the contrastive loss term only produces gradients with trivial magnitude. We use the notation $\mathtt{CE}(\cdot, \bm{y})$ to denote the sparse cross-entropy loss with a softmax attached, where $\bm{y}$ standards for the one-hot encoding label of the sample. Then the prototypical cross-entropy loss is computed as
\begin{equation}\label{eq:cel}
    L_{proto} = \sum\limits_{\bm{h} \in \{\mathcal{H}^p\cup \mathcal{H}^q\}} 
    \mathtt{CE}(g_p(\bm{h}), \bm{y})
\end{equation}
where $g_p(\cdot)$ is a linear projection head that maps the representations to the label.

Similar ideas have been explored in proxy learning works~\cite{yair2017no,kim2020proxy}.
However, our SPCL is distinct from them in terms of both the general motivation and the detailed implementation. 
1) Generally, they take the advantage of the proxy based loss for learning fine-grained semantic relations between data points to boost the speed of convergence. 
However, we utilize the Siamese-style metric loss to mitigate the semantic confusion of prototypes between false positives and the anchor. 
2) In details, they implement the Log-Sum-Exp as the max function to pull the anchor sample and most dissimiliar positive samples together and to push its most similar negative samples apart. In this work, we directly apply a prototypical cross entropy loss between the softmax output and the proxy prototype to guide the positive sample towards its prototypes with most similiar semantic relations. 
Our prototypical cross-entropy loss is much simpler than the Proxy-Anchor Loss, the variant of Proxy-NCA~\cite{yair2017no}.

\vspace{-1.5em}

\subsection{Overall Loss}
\vspace{-0.5em}
The overall loss is simply computed as the weighted summation of the three losses:
\begin{equation}\label{total_loss}
    L_{total} = \alpha L_{contra} + \beta L_{metric} +
    \gamma L_{proto}
\end{equation}
We carry out a series of ablation studies in Section \ref{sec:ab_each_loss} to explore the effects of each loss.

\vspace{-1.5em}
\section{Experimental Results}
\vspace{-0.5em}
\subsection{Dataset and Configurations}
\vspace{-0.5em}
We employ $3$ benchmark datasets for evaluating the classification performance: CIFAR-10/100~\cite{krizhevsky2009learning} and ImageNet-1K~\cite{imagenet_cvpr09}. For CIFAR-10/100 pre-training, we closely follow SimCLR~\cite{chen2020simple} and use the same data augmentation. 
For the encoder network ($f(\cdot)$) we experiment with the commonly used encoder architectures, ResNet-50. 
As the optimizer, we use LARS~\cite{yang2017scaling} with learning
rate of $1.0$ and weight decay of $10^{-6}$. 
We use linear warmup for the first 10 epochs, and decay the learning rate with the cosine decay schedule. 
We train at batch size $512$ for $1000$ epochs, using $512$ prototypes for clustering.\footnote{\label{note:sm}Detailed configurations of these experiments are provided in the supplementary materials.}

For ImageNet-1K, following~\cite{wang2020dense,chen2020mocov2}, we train it for $200$ epochs, using LARS with an initial learning rate of 1.0 and weight decay of $10^{-6}$. 
We train at a batch size of $1024$, and prototypes of $2048$ for clustering.
For clustering, we adopt the faiss library~\cite{JDH17faiss} for efficient $k$-means clustering, where it takes $6.5$ seconds per epoch during the pre-training. The clustering only takes less than $1,300$ seconds in total, which is fairly negligible compared to the whole training time (482 hours using 8 Tesla V100 GPUs).

\vspace{-0.5em}


\begin{table}[!htb]
	\caption{Comparison results of linear classification on ImageNet-1K. 
}
	\label{tab: cls_imagenet}
	\renewcommand\tabcolsep{3.0pt}
	\centering
	\scalebox{0.6}{
		\begin{tabular}{llccc}
			\toprule
			\textbf{Method} & \textbf{Arch.} & \textbf{Param.(M)} & \textbf{Batch} & \textbf{Top-1(\%)} \\
			\midrule
			\multicolumn{5}{l}{\textit{Instance-wise Contrastive Methods:}} \\
			CPC~\cite{oord2018representation} & ResNet-101 & $28$ & $2048$ & $48.70$ \\
			MoCo~\cite{he2019moco} & ResNet-50 & $24$ & $256$ & $60.60$ \\
			PIRL~\cite{misra2020self} & ResNet-50 & $24$ & $1024$ & $63.60$ \\
			CMC~\cite{tian2020contrastive} & ResNet-50+MLP$\{L, ab\}$ & $47$ & $128$ & $64.00$ \\
			CPCv2~\cite{henaff2019cpcv2} & ResNet-170 & $303$ & $512$ & $65.90$ \\
			AMDIM~\cite{bachman2019amdim} & Custom-ResNet & $192$ & $1008$ & $68.10$ \\
			CMC~\cite{tian2020contrastive} & ResNet-50 ($2\times$)+MLP & $192$ & $128$ & $68.40$ \\
			MoCo~\cite{he2019moco} & ResNet-50 ($4\times$) & $375$ & $256$ & $68.60$ \\
		    \multirow{3}{*}{SimCLR~\cite{chen2020simple}}  & ResNet-50+MLP & $28$ & $4096$ & $70.00$ \\
    		& ResNet-50 ($2\times$)+MLP & $98$ & $4096$ & $74.20$ \\
    		& ResNet-50 ($4\times$)+MLP & $379$ & $4096$ & $76.50$ \\    		MoCHi~\cite{kalantidis2020hard} & ResNet-50+MLP & $28$ & $512$ & $70.60$ \\
			MoCov2~\cite{chen2020mocov2} & ResNet-50+MLP & $28$ & $256$ & $71.10$ \\
			SimSiam~\cite{chen2021simsiam} & ResNet-50+MLP & $28$ & $256$ & $71.30$ \\
			SimCLRv2~\cite{chen2020big} & ResNet-50+MLP & $28$ & $4096$ & $71.70$ \\
		    BOYL~\cite{grill2020bootstrap} & ResNet-50+MLP & $35$ & $4096$ & $74.30$ \\
		    \midrule
            \multicolumn{5}{l}{\textit{Prototypical Contrastive Methods:}} \\
		    PCL~\cite{li2021prototypical} & ResNet-50 & $24$ & $256$ & $61.50$\\
		    PCLv2~\cite{li2021prototypical} & ResNet-50+MLP & $28$ & $256$ & $67.60$ \\
		    SwAV~\cite{caron2020unsupervised} & ResNet-50+MLP & $28$ & $4096$ & $75.30$ \\
			SPCL (ours) & ResNet-50+$\textrm{MLP}_{\times3}$ & $36$ & $1024$ & $\bm{77.68}$  \\
		    \midrule
		    \multicolumn{5}{l}{\textit{Supervised Methods:}} \\
			Supervised(\texttt{CE}) & ResNet-50 & $24$ & $256$ & $77.15$ \\
			SupCon~\cite{khosla2020sup} & ResNet-50+MLP & $28$ & $6144$ & $\underline{78.70}$ \\
			\bottomrule
			\end{tabular}}
			\vspace{-1em}
\end{table}
\vspace{-0.5em}

\subsection{Comparison with State-of-the-arts}
\vspace{-0.5em}
In Table \ref{tab: cls_imagenet}, we report the architecture, parameters, batch size, and top-1 accuracy of image classification on ImageNet-1K, where the MLP stands for the projection head. Since we employ $g_c, g_m, g_p$ for different proposals in SPCL, we denote it with $MLP_{\times3}$. It can be clearly seen that, our SPCL outperforms all the state-of-the-art methods CSL methods, except the SupCon which used the label information for end-to-end training. Especially, our method achieves $+0.53\%$ top-1 accuracy compared to the supervised model trained with the standard CE loss, shows better generalization in representation learning and semantic understanding. It is also worth mentioning that, we do not introduce a extreme large batch size setting as done in SimCLRv2, BOYL, and SwAV, nor have no dictionary/queue mechanism for tracking negative samples.

\vspace{-0.5em}

\subsection{Ablation Study}
\vspace{-0.5em}
\subsubsection{Batch Size and Number of Prototypes}\label{sec:ab_bs_np}
\vspace{-0.5em}
In this section, we study the effects of the \emph{batch size} and \emph{the number of prototypes}. Unless specified, we perform all ablation studies on the CIFAR-10 dataset. The batch size is one of the most important hyper-parameters in contrastive learning, former methods either unitize a genuine large batch size~\cite{chen2020big,grill2020bootstrap} or mechanisms that could simulate large batch sizes~\cite{chen2020mocov2,chen2021simsiam}. Here we also examine how the batch size affects the performance of our proposed method. The number of prototypes, on the other hand, is also crucial in the prototypical approaches~\cite{li2021prototypical,caron2020unsupervised}. We modify the number of prototypes in a wide range to investigate the robustness of our proposed method. Furthermore, the number of prototypes is not strictly specified, and we may stack multiple prototypical cross-entropy losses using different numbers of prototypes as in~\cite{li2021prototypical}. We also conduct experiments employing this setting. 

\begin{figure}[!htb]
\setlength{\abovecaptionskip}{-2em}
\setlength{\belowcaptionskip}{-1em}
\centering
   \includegraphics[width=0.9\linewidth]{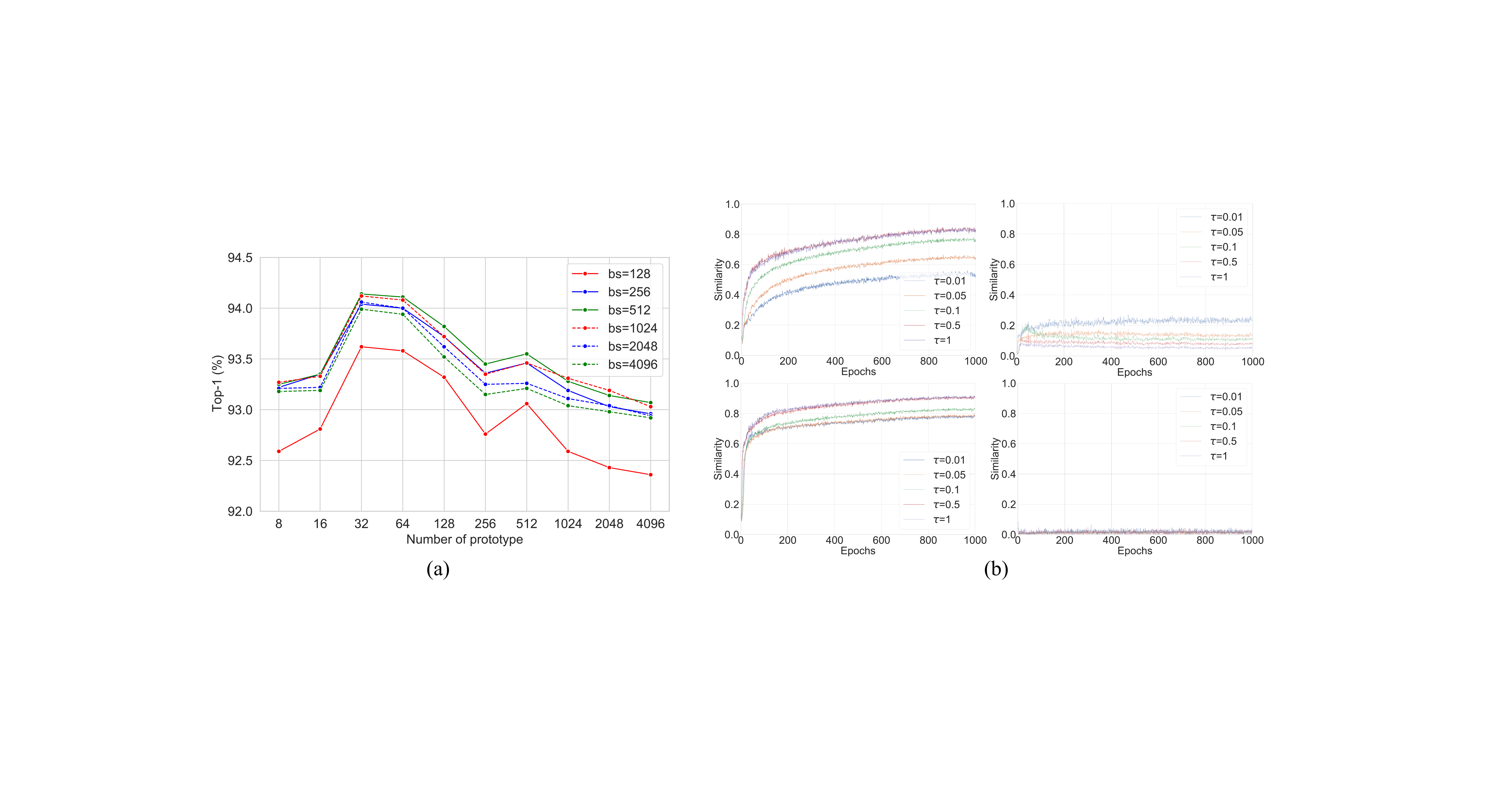}
   \vspace{0em}
   \caption{(a). Comparison of performance of top-1 accuracy w.r.t. various batch sizes and numbers of prototypes. (b). TP (\textbf{Left}) and FN (\textbf{Right}) similarity distance w.r.t. epoch at SimCLR (\textbf{Top}) and SPCL (\textbf{Bottom}) pre-training stage. Zoom in for a better view.}
\label{fig:ab_bs_proto_sim}
\end{figure}

Specifically, we vary the number of prototypes from $8$ to $4096$ and the batch size from $128$ to $4096$ to pre-train our SPCL on CIFAR-10 for linear evaluation. 
The experimental results are shown in Figure~\ref{fig:ab_bs_proto_sim} (a). 
From Figure~\ref{fig:ab_bs_proto_sim} (a), we can observe that the top-1 accuracy of linear evaluation does not have much change along with the number of the batch size, when it is sufficient (larger than 128 in this case). 
We conjecture that the regularization from the prototypical loss helps eliminates the demand for the large number of negative samples.

On the other hand, the accuracy is more-or-less (although not significantly) sensitive to the choice of numbers of prototypes. Intuitively, as long as the the number of prototypes does not match the number of classes in the down-stream task, noise in the semantic structure will be introduced.
In order to render this phoneme, we implement a symmetric prototypical cross-entropy loss, defined in~\cite{wang2019symmetric}, to overcome the noisy prototypes. 
We conduct experiments on CIFAR-100 using 512 prototypes, and the symmetric loss achieves better performance (74.04\%, 93.35\%) compared to asymmetric loss (73.97\%, 93.24\%) in terms of top-1 and top-5 accuracy.

\vspace{-1em}
\subsubsection{Effects of Losses}\label{sec:ab_each_loss}
\vspace{-0.5em}
In this section, we analyze three loss modules described in Section~\ref{sec:method}. 
Specifically, we compare the performance of top-1 accuracy, True Positive (TP), and False Negative (FN) distance by manipulating the weights of the loss modules (contrastive, metric, and prototype) proposed. 
The decision of whether a sample being positive or negative is based on the ground-truth of down-stream tasks. 
Concretely, we look at the ground-truth of an anchor sample, then mask all the negatively paired samples with the same label as the FN. 
The notation \textit{TP distance} refers to the similarity between an anchor image (\textbf{A}) and positive views (\textbf{TP}), and \textit{FN distance} is the similarity between A and FN, on the contrastive projection space, \textit{i.e.}. after $g_c(\cdot)$.
A larger TP distance means that positive views are similar to the anchor image, which will be beneficial for the positive sample terms in a contrastive learning loss. 
If FNs share semantic information with larger similarity to the anchor, the discriminability between positive and negative samples would be obscured, leading to pre-trained representations with low quality. 
On the other hand, FNs that share less similar semantic information with the anchor help the contrastive loss discriminate the difference between positive and negative samples during pre-training. 
With less FNs that share most similar semantic information with the anchor, pre-trained representations would be of high-quality and advantageous. 
\begin{table}[!htb]
\caption{Ablation study of each loss module (contrastive, metric, and prototype) on CIFAR-10 in terms of top-1 accuracy, True Positive (TP) and False Negative (FN) distance. CE and GT denotes the cross-entropy loss and the ground-truth, separately.}
\label{tab: ab_metric_cls_1}
\renewcommand\tabcolsep{6.0pt}
\centering
\scalebox{0.6}{
\begin{tabular}{cccccccc}
		\toprule
		method & contrastive($\alpha$) & metric($\beta$) & prototype($\gamma$) & top-1 (\%) & TP distance$(\uparrow)$ & FN distance$(\downarrow)$ \\
		\midrule
		Supervised (\texttt{CE}) & \xmark & \xmark & \cmark(GT) & $93.02\pm0.05$ & --- & ---\\
		SimCLR~\cite{chen2020simple} & \cmark & \xmark & \xmark & $92.01\pm0.13$ & $0.68\pm0.03$ & $0.19\pm0.03$\\
		SPCL & \cmark(1.0) & \cmark(1.0) & \cmark(GT) & $\bold{94.42}\pm0.09$ & $\bold{0.93}\pm0.02$ & $\bold{0.03}\pm0.04$ \\		SPCL & \cmark(1.0) & \cmark(1.0) & \cmark(1.0) & $\underline{94.12}\pm0.11$ & $\underline{0.85}\pm0.04$ & $\underline{0.11}\pm0.05$ \\
		SPCL & \cmark(1.0) & \cmark(1.0) & \xmark & $92.43\pm0.12$ & $0.76\pm0.04$ & $0.15\pm0.06$ \\
		SPCL & \cmark(1.0) & \xmark & \cmark(1.0) & $85.67\pm0.15$ & $0.58\pm0.02$ & $0.27\pm0.03$ \\
		SPCL & \xmark & \cmark(1.0) & \cmark(1.0) & $89.95\pm0.18$ & $0.61\pm0.03$ & $0.25\pm0.04$ \\ 
		\midrule
		SPCL & \cmark(\textcolor{blue}{0.1}) & \cmark(1.0) & \cmark(1.0) & $93.20\pm0.15$ & $0.81\pm0.05$ & $0.13\pm0.04$ \\
		SPCL & \cmark(1.0) & \cmark(\textcolor{blue}{0.1}) & \cmark(1.0) & $92.69\pm0.12$ & $0.78\pm0.04$ & $0.14\pm0.05$ \\
		SPCL & \cmark(1.0) & \cmark(1.0) & \cmark(\textcolor{blue}{0.1}) & $93.68\pm0.11$ & $0.83\pm0.03$ & $0.12\pm0.04$ \\
		\bottomrule
\end{tabular}
}
\end{table}

The results are reported in Table~\ref{tab: ab_metric_cls_1}. We either lower the importance of a loss or completely eliminate it. As can be seen, in the contrastive scene, using only the prototypical loss will not improve the performance over the baseline, while a small metric loss will significantly boost the performance in terms of both accuracy and the FN distance. This shows that the necessity of refining the quality of prototypes using the metric loss. On the other hand, removing the regularization from the prototypical loss also harms the performance. The pure clustering setting (that without the contrastive loss) also has degraded performance. 
Note that the FN distance in our proposed methods is much smaller than that in SimCLR, allows the model to build a better understanding of semantic information. 
Moreover, when employing the ground-truth label as the prototypes, the strong regularization effects eliminate the influence caused by FN samples, achieving the best top-1 accuracy.

In addition, we perform extensive experiments to analyze the effect of TP and FN distance on contrastive self-supervised learning. 
Specifically, we pre-train SimCLR and our proposed SPCL separately on the CIFAR-10 dataset by running $1000$ epochs for $10$ times. We calculate the TP/FN distance through the pre-training process by using different temperatures, as shown in Figure~\ref{fig:ab_bs_proto_sim} (b). 
As can be seen in SimCLR, the average FN distance become larger with the increase of running epochs. 
We can observe that those two distances are sensitive to the choice of the temperature. 
As the temperature $\tau$ increases, it gives a larger penalty to the samples closed to the anchor, which leads to the increase of the TP distance and the decrease of the FN distance. 
In the first 30 epochs, the mean FN distance arise to the maximum value, which influences the quality of the subsequently learned representations. 

Moreover, we report the TP and FN distance of our SPCL in two figures on the bottom row of Figure~\ref{fig:ab_bs_proto_sim} (b). 
It can be seen that the TP and FN distance become more robust to the choice of the temperature in the contrastive loss. 
Moreover, the TP distances in our SPCL are much higher than those in SimCLR, which means that our positives share strongly similar semantic information with the anchor. The FN distances in our SPCL are lower than those in SimCLR. In this case, the FN shares less similar semantic information with the anchor such that our SPCL mitigates the semantic confusion brought by FN.
In a nutshell, the increase of TP distances and the decrease of FN distances in our SPCL bring the true positives together and push semantically less similar false negatives apart in the pre-trained representations, boosting the performance of downstream tasks.

\vspace{-1.5em}
\section{Conclusion}
\vspace{-0.5em}
In this paper, we realize the possibility of \emph{perform supervised training in the unsupervised configuration}. We propose SPCL, a simple but effective Siamese-style prototypical contrastive self-supervised learning framework. Specifically, we group the embedded features into separate clusters by a prototypical cross-entropy loss and employ the Siamese-style metric loss to match intra-prototype features.
It is supposed to mitigate the semantically-similar confusion brought by false negatives in the contrastive loss term. We conduct extensive experiments on various benchmarks where the results demonstrate our method's effectiveness in improving the quality of visual representations for image classification.

\section*{Acknowledgement}

The work is partially supported by JSPS KAKENHI (Grant No. 20K19875), the National Natural Science Foundation of China (Grant No. 62006045).

\bibliography{reference}

\begin{thebibliography}{43}
\providecommand{\natexlab}[1]{#1}
\providecommand{\url}[1]{\texttt{#1}}
\expandafter\ifx\csname urlstyle\endcsname\relax
  \providecommand{\doi}[1]{doi: #1}\else
  \providecommand{\doi}{doi: \begingroup \urlstyle{rm}\Url}\fi

\bibitem[Arora et~al.(2019)Arora, Khandeparkar, Khodak, Plevrakis, and
  Saunshi]{arora2019theoretical}
Sanjeev Arora, Hrishikesh Khandeparkar, Mikhail Khodak, Orestis Plevrakis, and
  Nikunj Saunshi.
\newblock A theoretical analysis of contrastive unsupervised representation
  learning.
\newblock \emph{arXiv preprint arXiv:1902.09229}, 2019.

\bibitem[Bachman et~al.(2019)Bachman, Hjelm, and Buchwalter]{bachman2019amdim}
Philip Bachman, R~Devon Hjelm, and William Buchwalter.
\newblock Learning representations by maximizing mutual information across
  views.
\newblock In \emph{Proceedings of Advances in Neural Information Processing
  Systems (NeurIPS)}, 2019.

\bibitem[Becker and Hinton(1992)]{suzanna1992self-organizing}
Suzanna Becker and Geoffrey~E. Hinton.
\newblock Self-organizing neural network that discovers surfaces in random-dot
  stereograms.
\newblock \emph{Nature}, 355:\penalty0 161–163, 1992.

\bibitem[Caron et~al.(2018)Caron, Bojanowski, Joulin, and Douze]{caron2018deep}
Mathilde Caron, Piotr Bojanowski, Armand Joulin, and Matthijs Douze.
\newblock Deep clustering for unsupervised learning of visual features.
\newblock In \emph{Proceedings of European Conference on Computer Vision},
  2018.

\bibitem[Caron et~al.(2019)Caron, Bojanowski, Mairal, and
  Joulin]{caron2019unsupervised}
Mathilde Caron, Piotr Bojanowski, Julien Mairal, and Armand Joulin.
\newblock Unsupervised pre-training of image features on non-curated data.
\newblock In \emph{Proceedings of the International Conference on Computer
  Vision (ICCV)}, pages 2959--2968, 2019.

\bibitem[Caron et~al.(2020)Caron, Misra, Mairal, Goyal, Bojanowski, and
  Joulin]{caron2020unsupervised}
Mathilde Caron, Ishan Misra, Julien Mairal, Priya Goyal, Piotr Bojanowski, and
  Armand Joulin.
\newblock Unsupervised learning of visual features by contrasting cluster
  assignments.
\newblock In \emph{Proceedings of Advances in Neural Information Processing
  Systems (NeurIPS)}, 2020.

\bibitem[Chen et~al.(2020{\natexlab{a}})Chen, Kornblith, Norouzi, and
  Hinton]{chen2020simple}
Ting Chen, Simon Kornblith, Mohammad Norouzi, and Geoffrey Hinton.
\newblock A simple framework for contrastive learning of visual
  representations.
\newblock In \emph{Proceedings of International Conference on Machine
  Learning}, 2020{\natexlab{a}}.

\bibitem[Chen et~al.(2020{\natexlab{b}})Chen, Kornblith, Swersky, Norouzi, and
  Hinton]{chen2020big}
Ting Chen, Simon Kornblith, Kevin Swersky, Mohammad Norouzi, and Geoffrey
  Hinton.
\newblock Big self-supervised models are strong semi-supervised learners.
\newblock In \emph{Proceedings of Advances in Neural Information Processing
  Systems (NeurIPS)}, 2020{\natexlab{b}}.

\bibitem[Chen and He(2021)]{chen2021simsiam}
Xinlei Chen and Kaiming He.
\newblock Exploring simple siamese representation learning.
\newblock In \emph{Proceedings of the IEEE/CVF Conference on Computer Vision
  and Pattern Recognition (CVPR)}, pages 15750--15758, 2021.

\bibitem[Chen et~al.(2020{\natexlab{c}})Chen, Fan, Girshick, and
  He]{chen2020mocov2}
Xinlei Chen, Haoqi Fan, Ross Girshick, and Kaiming He.
\newblock Improved baselines with momentum contrastive learning.
\newblock \emph{arXiv preprint arXiv:2003.04297}, 2020{\natexlab{c}}.

\bibitem[Deng et~al.(2009)Deng, Dong, Socher, Li, Li, and
  Fei-Fei]{imagenet_cvpr09}
Jia Deng, Wei Dong, Richard Socher, Li-Jia. Li, Kai Li, and Li~Fei-Fei.
\newblock {ImageNet: A Large-Scale Hierarchical Image Database}.
\newblock In \emph{Proceedings of the IEEE/CVF Conference on Computer Vision
  and Pattern Recognition (CVPR)}, pages 248--255, 2009.

\bibitem[Dosovitskiy et~al.(2014)Dosovitskiy, Springenberg, Riedmiller, and
  Brox]{dosovitskiy2014discriminative}
Alexey Dosovitskiy, Jost~Tobias Springenberg, Martin Riedmiller, and Thomas
  Brox.
\newblock Discriminative unsupervised feature learning with convolutional
  neural networks.
\newblock In \emph{Proceedings of Advances in Neural Information Processing
  Systems (NeurIPS)}, 2014.

\bibitem[F\"{o}ldi\'{a}k(1991)]{peter1991learning}
Peter F\"{o}ldi\'{a}k.
\newblock Learning invariance from transformation sequences.
\newblock \emph{Neural Computation}, 3(2):\penalty0 194--200, 1991.

\bibitem[Grill et~al.(2020)Grill, Strub, Altch\'{e}, Tallec, Richemond,
  Buchatskaya, Doersch, Avila~Pires, Guo, Gheshlaghi~Azar, Piot, kavukcuoglu,
  Munos, and Valko]{grill2020bootstrap}
Jean-Bastien Grill, Florian Strub, Florent Altch\'{e}, Corentin Tallec, Pierre
  Richemond, Elena Buchatskaya, Carl Doersch, Bernardo Avila~Pires, Zhaohan
  Guo, Mohammad Gheshlaghi~Azar, Bilal Piot, koray kavukcuoglu, Remi Munos, and
  Michal Valko.
\newblock Bootstrap your own latent - a new approach to self-supervised
  learning.
\newblock In \emph{Proceedings of Advances in Neural Information Processing
  Systems (NeurIPS)}, 2020.

\bibitem[Hadsell et~al.(2006)Hadsell, Chopra, and
  LeCun]{hadsell2006dimensionality}
Raia Hadsell, Sumit Chopra, and Yann LeCun.
\newblock Dimensionality reduction by learning an invariant mapping.
\newblock In \emph{Proceedings of the IEEE/CVF Conference on Computer Vision
  and Pattern Recognition (CVPR)}, pages 1735--1742, 2006.

\bibitem[He et~al.(2020)He, Fan, Wu, Xie, and Girshick]{he2019moco}
Kaiming He, Haoqi Fan, Yuxin Wu, Saining Xie, and Ross Girshick.
\newblock Momentum contrast for unsupervised visual representation learning.
\newblock In \emph{Proceedings of the IEEE/CVF Conference on Computer Vision
  and Pattern Recognition (CVPR)}, pages 9729--9738, 2020.

\bibitem[H{\'{e}}naff et~al.(2019)H{\'{e}}naff, Srinivas, Fauw, Razavi,
  Doersch, Eslami, and van~den Oord]{henaff2019cpcv2}
Olivier~J. H{\'{e}}naff, Aravind Srinivas, Jeffrey~De Fauw, Ali Razavi, Carl
  Doersch, S.~M.~Ali Eslami, and A{\"{a}}ron van~den Oord.
\newblock Data-efficient image recognition with contrastive predictive coding.
\newblock \emph{arXiv preprint arXiv:1905.09272}, 2019.

\bibitem[Johnson et~al.(2017)Johnson, Douze, and J{\'e}gou]{JDH17faiss}
Jeff Johnson, Matthijs Douze, and Herv{\'e} J{\'e}gou.
\newblock Billion-scale similarity search with gpus.
\newblock \emph{arXiv preprint arXiv:1702.08734}, 2017.

\bibitem[Kalantidis et~al.(2020)Kalantidis, Sariyildiz, Pion, Weinzaepfel, and
  Larlus]{kalantidis2020hard}
Yannis Kalantidis, Mert~Bulent Sariyildiz, Noe Pion, Philippe Weinzaepfel, and
  Diane Larlus.
\newblock Hard negative mixing for contrastive learning.
\newblock In \emph{Proceedings of Advances in Neural Information Processing
  Systems (NeurIPS)}, 2020.

\bibitem[Khosla et~al.(2020)Khosla, Teterwak, Wang, Sarna, Tian, Isola,
  Maschinot, Liu, and Krishnan]{khosla2020sup}
Prannay Khosla, Piotr Teterwak, Chen Wang, Aaron Sarna, Yonglong Tian, Phillip
  Isola, Aaron Maschinot, Ce~Liu, and Dilip Krishnan.
\newblock Supervised contrastive learning.
\newblock In \emph{Proceedings of Advances in Neural Information Processing
  Systems (NeurIPS)}, 2020.

\bibitem[Kim et~al.(2020)Kim, Kim, Cho, and Kwak]{kim2020proxy}
Sungyeon Kim, Dongwon Kim, Minsu Cho, and Suha Kwak.
\newblock Proxy anchor loss for deep metric learning.
\newblock In \emph{Proceedings of the IEEE/CVF Conference on Computer Vision
  and Pattern Recognition (CVPR)}, pages 3431--3440, 2020.

\bibitem[Koch et~al.(2015)Koch, Zemel, and Salakhutdinov]{koch2015siamese}
Gregory Koch, Richard Zemel, and Ruslan Salakhutdinov.
\newblock Siamese neural networks for one-shot image recognition.
\newblock In \emph{Proceedings of International Conference on Machine Learning
  (ICML) Deep Learning Workshop}, 2015.

\bibitem[Krizhevsky(2009)]{krizhevsky2009learning}
Alex Krizhevsky.
\newblock Learning multiple layers of features from tiny images.
\newblock 2009.

\bibitem[Li et~al.(2021)Li, Zhou, Xiong, and Hoi]{li2021prototypical}
Junnan Li, Pan Zhou, Caiming Xiong, and Steven Hoi.
\newblock Prototypical contrastive learning of unsupervised representations.
\newblock In \emph{Proceedings of International Conference on Learning
  Representations (ICLR)}, 2021.

\bibitem[Misra and Maaten(2020)]{misra2020self}
Ishan Misra and Laurens van~der Maaten.
\newblock Self-supervised learning of pretext-invariant representations.
\newblock In \emph{Proceedings of the IEEE/CVF Conference on Computer Vision
  and Pattern Recognition (CVPR)}, pages 6707--6717, 2020.

\bibitem[Movshovitz-Attias et~al.(2017)Movshovitz-Attias, Toshev, Leung, Ioffe,
  and Singh]{yair2017no}
Yair Movshovitz-Attias, Alexander Toshev, Thomas~K. Leung, Sergey Ioffe, and
  Saurabh Singh.
\newblock No fuss distance metric learning using proxies.
\newblock In \emph{Proceedings of the International Conference on Computer
  Vision (ICCV)}, pages 360--368, 2017.

\bibitem[Noroozi and Favaro(2016)]{noroozi2016unsupervised}
Mehdi Noroozi and Paolo Favaro.
\newblock Unsupervised learning of visual representations by solving jigsaw
  puzzles.
\newblock In \emph{Proceedings of European Conference on Computer Vision
  (ECCV)}, 2016.

\bibitem[Oord et~al.(2018)Oord, Li, and Vinyals]{oord2018representation}
Aaron van~den Oord, Yazhe Li, and Oriol Vinyals.
\newblock Representation learning with contrastive predictive coding.
\newblock \emph{arXiv preprint arXiv:1807.03748}, 2018.

\bibitem[Pathak et~al.(2016)Pathak, Kr{\"{a}}henb{\"{u}}hl, Donahue, Darrell,
  and Efros]{deepak2016context}
Deepak Pathak, Philipp Kr{\"{a}}henb{\"{u}}hl, Jeff Donahue, Trevor Darrell,
  and Alexei~A. Efros.
\newblock Context encoders: Feature learning by inpainting.
\newblock In \emph{Proceedings of the IEEE/CVF Conference on Computer Vision
  and Pattern Recognition (CVPR)}, pages 2536--2544, 2016.

\bibitem[Poole et~al.(2019)Poole, Ozair, Van Den~Oord, Alemi, and
  Tucker]{poole2019variational}
Ben Poole, Sherjil Ozair, Aaron Van Den~Oord, Alex Alemi, and George Tucker.
\newblock On variational bounds of mutual information.
\newblock In \emph{Proceedings of International Conference on Machine Learning
  (ICML)}, pages 5171--5180, 2019.

\bibitem[Roh et~al.(2021)Roh, Shin, Kim, and Kim]{byungseok2021spatial}
Byungseok Roh, Wuhyun Shin, Ildoo Kim, and Sungwoong Kim.
\newblock Spatially consistent representation learning.
\newblock In \emph{Proceedings of the IEEE/CVF Conference on Computer Vision
  and Pattern Recognition (CVPR)}, pages 1144--1153, 2021.

\bibitem[Sa(1993)]{virginia1993learning}
Virginia R.~De Sa.
\newblock Learning classification with unlabeled data.
\newblock In \emph{Proceedings of Advances in Neural Information Processing
  Systems (NeurIPS)}, page 112–119, 1993.

\bibitem[Sohn(2016)]{sohn2016improved}
Kihyuk Sohn.
\newblock Improved deep metric learning with multi-class n-pair loss objective.
\newblock In \emph{Proceedings of Advances in Neural Information Processing
  Systems (NeurIPS)}, 2016.

\bibitem[Tian et~al.(2020)Tian, Krishnan, and Isola]{tian2020contrastive}
Yonglong Tian, Dilip Krishnan, and Phillip Isola.
\newblock Contrastive multiview coding.
\newblock In \emph{Proceedings of European Conference on Computer Vision
  (ECCV)}, 2020.

\bibitem[Tschannen et~al.(2019)Tschannen, Djolonga, Rubenstein, Gelly, and
  Lucic]{tschannen2019mutual}
Michael Tschannen, Josip Djolonga, Paul~K Rubenstein, Sylvain Gelly, and Mario
  Lucic.
\newblock On mutual information maximization for representation learning.
\newblock \emph{arXiv preprint arXiv:1907.13625}, 2019.

\bibitem[Wang et~al.(2021)Wang, Zhang, Shen, Kong, and Li]{wang2020dense}
Xinlong Wang, Rufeng Zhang, Chunhua Shen, Tao Kong, and Lei Li.
\newblock Dense contrastive learning for self-supervised visual pre-training.
\newblock In \emph{Proceedings of the IEEE/CVF Conference on Computer Vision
  and Pattern Recognition (CVPR)}, pages 3024--3033, 2021.

\bibitem[Wang et~al.(2019)Wang, Ma, Chen, Luo, Yi, and
  Bailey]{wang2019symmetric}
Yisen Wang, Xingjun Ma, Zaiyi Chen, Yuan Luo, Jinfeng Yi, and James Bailey.
\newblock Symmetric cross entropy for robust learning with noisy labels.
\newblock In \emph{Proceedings of the International Conference on Computer
  Vision (ICCV)}, pages 322--330, 2019.

\bibitem[Wu et~al.(2018)Wu, Xiong, Yu, and Lin]{wu2018unsupervised}
Zhirong Wu, Yuanjun Xiong, Stella Yu, and Dahua Lin.
\newblock Unsupervised feature learning via non-parametric instance
  discrimination.
\newblock In \emph{Proceedings of the IEEE/CVF Conference on Computer Vision
  and Pattern Recognition (CVPR)}, pages 3733--3742, 2018.

\bibitem[Xiao et~al.(2020)Xiao, Wang, Efros, and Darrell]{xiao2020contrastive}
Tete Xiao, Xiaolong Wang, Alexei~A. Efros, and Trevor Darrell.
\newblock What should not be contrastive in contrastive learning.
\newblock \emph{arXiv preprint arXiv:2008.05659}, 2020.

\bibitem[Yamaguchi et~al.(2019)Yamaguchi, Kanai, Shioda, and
  Takeda]{yamaguchi2019multiple}
Shin'ya Yamaguchi, Sekitoshi Kanai, Tetsuya Shioda, and Shoichiro Takeda.
\newblock Multiple pretext-task for self-supervised learning via mixing
  multiple image transformations.
\newblock \emph{arXiv preprint arXiv:1912.11603}, 2019.

\bibitem[You et~al.(2017)You, Gitman, and Ginsburg]{yang2017scaling}
Yang You, Igor Gitman, and Boris Ginsburg.
\newblock Scaling {SGD} batch size to 32k for imagenet training.
\newblock \emph{arXiv preprint arXiv:1708.03888}, 2017.

\bibitem[Zhan et~al.(2020)Zhan, Xie, Liu, Ong, and Loy]{zhan2020online}
Xiaohang Zhan, Jiahao Xie, Ziwei Liu, Yew-Soon Ong, and Chen~Change Loy.
\newblock Online deep clustering for unsupervised representation learning.
\newblock In \emph{Proceedings of the IEEE/CVF Conference on Computer Vision
  and Pattern Recognition (CVPR)}, pages 6688--6697, 2020.

\bibitem[Zhang et~al.(2016)Zhang, Isola, and Efros]{zhang2016colorful}
Richard Zhang, Phillip Isola, and Alexei~A Efros.
\newblock Colorful image colorization.
\newblock In \emph{Proceedings of European Conference on Computer Vision
  (ECCV)}, 2016.

\end{thebibliography}

\end{document}